%% file: main_arxiv.tex
\documentclass[journal]{IEEEtran}
\ifCLASSINFOpdf

\else

\fi
\usepackage{amsmath,amsfonts}
\usepackage{algorithmic}
\usepackage{algorithm}
\usepackage{array}
\usepackage[caption=false,font=normalsize,labelfont=sf,textfont=sf]{subfig}
\usepackage{capt-of}
\usepackage{textcomp}
\usepackage{stfloats}
\usepackage{url}
\usepackage{verbatim}
\usepackage{graphicx}
\usepackage{cite}

\usepackage{multirow} 
\usepackage{booktabs}
\usepackage{hyperref}  
\usepackage{hyperref}
\usepackage{amsmath}
\usepackage{amssymb}
\usepackage{mathtools}
\usepackage{pifont}
\usepackage{booktabs}   
\usepackage{multirow}   
\usepackage{graphicx}   
\usepackage{colortbl}
\definecolor{headergray}{RGB}{242, 242, 242} 
\definecolor{ourblue}{RGB}{230, 240, 255}   
\definecolor{tablegray}{gray}{0.92}
\usepackage[capitalize,noabbrev]{cleveref}
\usepackage[table,dvipsnames]{xcolor} 

\usepackage{microtype}

\begin{document}

\title{Visual Geometry Foundation-Aware Gaussians for Single-Frame Surround-View Driving Reconstruction}



\author{
Junhong Lin, Jinlong Wang, Xianda Guo, Yanlun Peng, Wei Zheng, Guoqing Liu, 
Hanli Wang,~\IEEEmembership{Senior Member, IEEE}, Tiesong Zhao,~\IEEEmembership{Senior Member, IEEE}, and Wei Gao,~\IEEEmembership{Senior Member, IEEE}
\thanks{
This work was supported by National Science and Technology Major Project (2024ZD01NL00101), Natural Science Foundation of China (62271013), Guangdong Provincial Key Laboratory of Ultra High Definition Immersive Media Technology (2024B1212010006), Guangdong Province Pearl River Talent Program (2021QN020708), Guangdong Basic and Applied Basic Research Foundation (2024A1515010155), Shenzhen Science and Technology Program (JCYJ20240813160202004, JCYJ20230807120808017, SYSPG20241211173440004), Shenzhen Fundamental Research Program (GXWD20201231165807007-20200806163656003), and financially supported for Outstanding Talents Training Fund in Shenzhen. (\textit{Corresponding author: Wei Gao})}

\thanks{Junhong Lin, Jinlong Wang and Wei Gao are with Guangdong Provincial Key Laboratory of Ultra High Definition Immersive Media Technology, School of Electronic and Computer Engineering, Peking University, Shenzhen 518055, China, and Wei Gao is also with Peng Cheng Laboratory, Shenzhen 518066, China. (e-mail: \{jhlin42in, jinlongw128\}@gmail.com, gaowei262@pku.edu.cn).

Xianda Guo is with Wuhan University, Wuhan, China. Yanlun Peng is with Great Wall Motor, China. Wei Zheng, Guoqing Liu is with Minieye Corporation, Shenzhen 518055, China.
Hanli Wang is with Tongji University, Shanghai, 200100, China. 
Tiesong Zhao is with Fuzhou University, Fuzhou, 350108, China.
(e-mail: xianda\_guo@163.com, yanlunpeng@gwm.cn, \{zhengwei, guoqing\}@minieye.cc, hanliwang@tongji.edu.cn, t.zhao@fzu.edu.cn).
}

\thanks{Our code will be available at: https://github.com/JHLin42in/VGGD.}

}


\maketitle

\begin{abstract}
Single-frame surround-view reconstruction faces severe geometric instability and rendering artifacts due to minimal inter-camera overlap. While existing methods rely on complex decoders or auxiliary cues, they remain bottlenecked by the weak geometric capacity of upstream features. We argue that leveraging pretrained visual geometry priors strengthens upstream representations and alleviates the geometric ambiguity in sparse surround views. To this end, we propose \textbf{VGGD}, a visual geometry foundation-aware 3D Gaussian Splatting framework for feed-forward surround-view driving reconstruction, which shifts geometric modeling to the frontend and adapts foundation priors to the driving camera setting. First, VGGD leverages VGGT to provide transferable multi-view geometric prior tokens. Next, we introduce a \emph{Dual-Path Neck} to decouple geometry-consistent and appearance-aware representations, improving appearance completion in weakly observed regions. We further apply \emph{Scale Warmup} to stabilize early geometry learning and suppress scale drift under ego-pose changes. Finally, we use a hybrid pixel--volume Gaussian decoder to produce a renderable 3D Gaussian scene for novel-view synthesis. Experiments on the nuScenes single-frame benchmark show that VGGD achieves the best overall rendering quality among the compared methods and improves relative geometric consistency.

\end{abstract}


\IEEEpeerreviewmaketitle

\section{Introduction}
\label{sec:intro}

High-fidelity 3D reconstruction from sparse surround-view cameras is a fundamental capability in autonomous driving, enabling closed-loop simulation, controllable data generation, and high-definition mapping~\cite{wei2025omniscene,tian2024drivingforward,yang2025storm,ye2026autodrivep3}.
In practice, vehicles are equipped with only a few cameras, typically six surround views, with very limited overlap between adjacent fields of view~\cite{caesar2020nuscenes,wei2025omniscene}.
Unlike object-centric or indoor reconstruction, driving scenes are spatially extensive and contain large depth variations, thin structures, dynamic agents, and frequent occlusions, which further amplify the ambiguity of sparse-view geometry.
In the single-frame feed-forward setting, the model must recover a scene within a spatial extent around the ego vehicle from one sparse surround capture and synthesize novel views under subsequent ego-pose changes~\cite{wei2025omniscene,tian2024drivingforward,lin2025vgd}.
Moreover, this feed-forward formulation leaves no opportunity for per-scene optimization, requiring the model to infer metric structure and appearance completion in a single forward pass.
With scarce cross-view correspondences, geometric inference is strongly ill-posed and easily under-constrained~\cite{wei2025omniscene,charatan2024pixelsplat,chen2024mvsplat}.
Small geometric inconsistencies become pronounced under meter-level viewpoint shifts, leading to structural misalignment and rendering artifacts.
This makes high-quality reconstruction in real driving scenes particularly challenging.

\begin{figure}[t]
  \centering
  \includegraphics[width=1.0\linewidth]{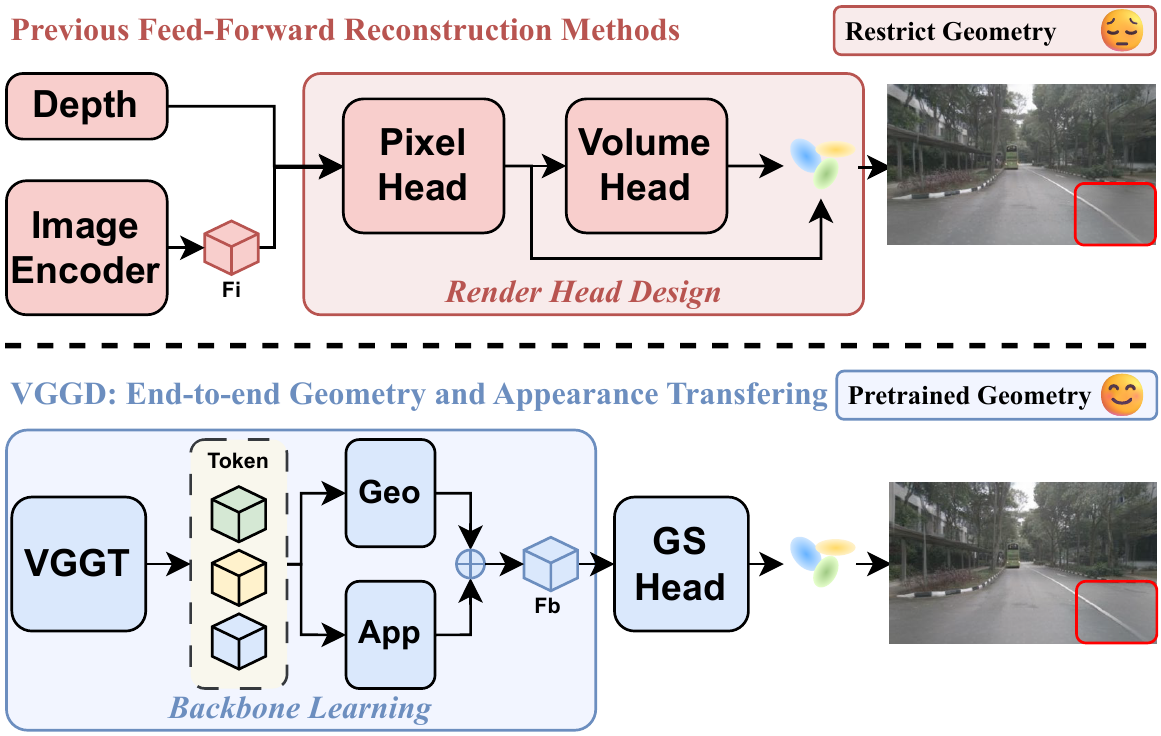}
  \caption{\textbf{Adapting geometry foundation priors for sparse surround-view reconstruction.} Under minimal inter-camera overlap, previous feed-forward pipelines rely on weak geometric cues and place most burden on head design, often causing scale drift and structural inconsistency that appear as distortions and artifacts in novel-view rendering (top). VGGD (bottom) instead leverages a pretrained geometry foundation model (VGGT) and introduces \emph{Scale Warmup} and a \emph{Dual-Path Neck} to stabilize geometry and enhance appearance features, yielding more faithful 3DGS reconstructions.}
  \label{fig:teaser}
\end{figure}

Recent feed-forward methods have largely advanced this task through renderable representation design and specialized decoding heads, with 3D Gaussian Splatting widely adopted for efficient differentiable rendering~\cite{kerbl20233dgs,tian2024drivingforward,wei2025omniscene,charatan2024pixelsplat,chen2024mvsplat}.
However, 3DGS still depends on reliable 3D support, and decoder design alone cannot resolve missing geometric constraints.
As highlighted in Figure~\ref{fig:teaser} (top), these pipelines typically rely on weak or externally provided geometric cues and place most burden on head design to recover geometry and appearance~\cite{wei2025omniscene,tian2024drivingforward,yang2024depthanythingv2,hu2024metric3dv2}.
As a result, scale, visibility, or alignment errors can directly lead to misplaced Gaussians and unstable rendering.
While such strategies improve results, geometry remains fragile under minimal overlap, and scale drift and view-inconsistent structure still emerge and propagate to novel-view rendering~\cite{wei2025omniscene}.
Meanwhile, appearance completion in occluded or weakly observed regions is often limited by geometry-biased features, resulting in visible artifacts~\cite{charatan2024pixelsplat,chen2024mvsplat}.

Motivated by this analysis, our key insight is to shift the modeling focus from head-centric geometry estimation to transferable upstream geometric priors~\cite{wang2025vggt,wang2024dust3r,cabon2024mast3r}. As illustrated in Figure~\ref{fig:teaser}, we leverage a pretrained geometry foundation model to provide structural representations that directly guide Gaussian prediction, rather than serving merely as auxiliary depth cues~\cite{wang2025vggt,lin2025vgd}. 
However, open-domain geometry priors may not align with surround-view driving cameras and can exhibit depth-scale instability under ego-pose changes~\cite{hu2024metric3dv2,wei2025omniscene}. Geometry-dominant representations may also suppress semantic and high-frequency cues needed for faithful appearance completion~\cite{wei2025omniscene}.
These observations motivate driving-oriented adaptation for stable geometry and preserved appearance.

To address these challenges, we propose \textbf{VGGD}, a geometry foundation-aware Gaussian Splatting framework for single-frame surround-view driving reconstruction.
Instead of relying primarily on increasingly specialized Gaussian heads, VGGD shifts geometric reasoning to a geometry foundation frontend, which extracts transferable multi-view structural priors from sparse surround images.
To adapt these priors to the driving-camera setting, we design a \emph{Dual-Path Neck} that separates geometry-consistent features for stable 3D support from appearance-aware features for texture and semantic completion.
To mitigate scale instability under ego-pose changes, we further introduce \emph{Scale Warmup}, which transiently anchors early geometry learning and then removes this constraint to allow end-to-end rendering-driven refinement.
Finally, we employ a standard pixel--volume 3DGS decoder to convert the adapted features and geometry into a renderable Gaussian scene without introducing additional decoder complexity.
The main contributions of our work are summarized as follows:

\begin{itemize}
   
    \item We propose \textbf{VGGD}, a visual geometry foundation-aware 3DGS framework for feed-forward single-frame sparse surround-view driving reconstruction, which shifts the bottleneck from head-centric decoding to geometry prior enhanced frontend representations.
    
    \item We introduce a driving-oriented architecture that integrates \emph{Dual-Path Neck} to model geometry-consistent and appearance-aware features for reliable rendering, and \emph{Scale Warmup} to stabilize early scale learning.
    
    \item Extensive experiments on the nuScenes ego-centric single-frame reconstruction benchmark demonstrate that VGGD achieves state-of-the-art novel-view synthesis quality and improved geometric consistency.

\end{itemize}

\section{Related Work}
\label{sec:related_work}

\subsection{Pretrained Visual and Geometric Priors}
\label{subsec:rw_pretrain}

\subsubsection{Visual foundation features}
Large-scale pretraining provides transferable visual representations that are widely used as backbones, including ViT-based self-supervised models (DINO/DINOv2)~\cite{caron2021dino,oquab2023dinov2} and masked-image modeling (MAE)~\cite{he2022mae}, as well as language-supervised features (CLIP)~\cite{radford2021clip}.
Such features substantially improve robustness and generalization in downstream perception tasks, and are often adopted in reconstruction pipelines as strong semantic encoders.
However, they are not explicitly trained to provide metrically consistent geometry or cross-view structural constraints under extremely low-overlap multi-view inputs, which is crucial for single-frame surround-view reconstruction.

\subsubsection{Geometric priors and geometry foundation models}
Dense-prediction depth models~\cite{ranftl2021dpt} and modern depth foundation priors ~\cite{yang2024depthanything,yang2024depthanythingv2,hu2024metric3dv2} offer convenient geometric cues and are frequently used to stabilize reconstruction under sparse observations.
These cues also benefit sparse-view neural rendering by providing robust depth regularization signals~\cite{wang2023sparsenerf,li2024dngaussian}.
While these priors improve single-view depth plausibility, cross-view consistency and metric alignment remain challenging in the minimal-overlap surround-view regime, often leading to scale drift or view-wise depth discrepancies.
Recently, geometry foundation models~\cite{wang2024dust3r,cabon2024mast3r,wang2025vggt} encode richer multi-view structure and enable feed-forward geometric reasoning beyond depth-only supervision.
In contrast, we emphasize a visual geometry foundation-aware pipeline for the single-frame setting.
This trend motivates combining pretrained geometry priors with driving-oriented adaptation for metric-scale single-frame reconstruction and high-fidelity rendering.

\subsection{Feed-forward Reconstruction for Autonomous Driving}
\label{subsec:rw_driving_ff}

\subsubsection{Generalizable feed-forward 3DGS from sparse views}
Generalizable 3D Gaussian Splatting (3DGS) methods regress Gaussians in a feed-forward manner to avoid per-scene optimization (e.g., NeRF)~\cite{mildenhall2020nerf,kerbl20233dgs}.
Representative works include single- or few-view Gaussian regression and sparse-view reconstruction pipelines~\cite{charatan2024pixelsplat,chen2024mvsplat}, which demonstrate the practicality of predicting explicit, renderable Gaussians from limited images.
Few-shot splatting-based optimization is explored in FSGS~\cite{zhu2023fsgs}.
Recent works further improve 3DGS with alias-free filtering, surface-aware parameterization, surface extraction, and SLAM~\cite{guedon2024sugar,wu2026fieldgs,matsuki2024gaussianslam,zhao2026msasplatting}.
These methods highlight the importance of reliable geometry cues (e.g., multi-view constraints or depth priors) when observations are sparse, and provide strong baselines for generalizable reconstruction.
Dynamic extensions include 4D Gaussian Splatting~\cite{wu2024_4dgs}, Spacetime Gaussian Feature Splatting~\cite{li2024spacetimegs}, and SplatFields~\cite{liang2026_4dgstream}.
Nevertheless, most are developed for generic sparse-view settings with moderate overlap, and do not directly address the extreme low-overlap surround-view configuration and meter-level ego-pose changes required in driving benchmarks.

\subsubsection{Surround-view driving reconstruction}
Single-frame surround-view benchmarks for driving scenes (e.g., nuScenes~\cite{caesar2020nuscenes}) explicitly expose the extremely low-overlap regime and the need for metric-consistent rendering under ego-pose changes.
DrivingForward~\cite{tian2024drivingforward} and Omni-Scene~\cite{wei2025omniscene} improve robustness by incorporating auxiliary geometry cues (e.g., metric depth priors) and hybrid pixel--volume Gaussian parameterizations to better handle occlusions and frustum truncations.
VGD~\cite{lin2025vgd} further explores distilling geometric priors from pretrained geometry models, while UniSplat~\cite{shi2025unisplat} extends feed-forward reconstruction towards unified spatio-temporal fusion for dynamic driving scenes. 
Beyond these, Some works explore surround-view driving representations for occupancy prediction and world modeling~\cite{lin2026vgocc, li2025omninwm, zhu2026gem}.
We adapt pretrained geometry priors to metric driving scale and enhance appearance features, complementing decoder-centric advances to improve geometric stability and rendering fidelity under minimal overlap.


\section{Methodology}
\label{sec:method}

\subsection{Task formulation}

Single-frame surround-view driving reconstruction aims to recover 3D scenes from sparse input and synthesize novel views under ego-pose changes.
We follow the single-frame surround-view setting as follows: given $V{=}6$ synchronized images
$\mathcal{I}_t=\{I_t^v\}_{v=1}^{V}$ and camera parameters $\{\mathbf{K}^v,\mathbf{T}_t^v\}_{v=1}^{V}$, the goal is scene-level 3D reconstruction that supports novel-view synthesis under ego-pose changes.
Given a latent scene representation $S_t$ inferred from $\mathcal{I}_t$, target views at displaced ego poses $\tau\in\mathcal{T}$ are rendered as $\hat I_\tau^v=\mathcal{R}(S_t;\mathbf{K}^v,\mathbf{T}_\tau^v)$.
In this task, $S_t$ is determined by two key latent variables, a camera-consistent 3D structure $\mathcal{X}_t$ and view-dependent appearance $\mathcal{A}_t$.
Accordingly, we write the rendering objective as:
\begin{equation}
\hat I_\tau^v \;=\; \mathcal{R}\!\big(\mathcal{X}_t,\mathcal{A}_t;\mathbf{K}^v,\mathbf{T}_\tau^v\big),
\qquad \tau\in\mathcal{T},\ v\in[1,V].
\label{eq:task_render}
\end{equation}

Thus, solving the task reduces to inferring $(\mathcal{X}_t,\mathcal{A}_t)$ from a single sparse surround-view capture, where minimal inter-camera overlap makes estimating $\mathcal{X}_t$ particularly ill-conditioned and critical for stable novel-view rendering.
Here, $\mathcal{X}_t$ determines the 3D support, visibility, and reprojection consistency under ego-pose changes, while $\mathcal{A}_t$ models the radiance and appearance attached to that support.
Therefore, structural errors such as depth-scale bias or cross-view misalignment can be amplified into persistent artifacts in target views.
The central challenge is to recover a camera-consistent structure while preserving faithful appearance from single-frame sparse surround capture.

\subsection{Limitations of existing methods}

Given the central role of $\mathcal{X}_t$, we next examine how existing feed-forward 3DGS pipelines obtain this structure.
Most feed-forward pipelines instantiate $S_t$ with a 3D Gaussian Splatting (3DGS) representation and predict it using task-specific decoding heads.
A typical design first estimates a geometry proxy (e.g., per-view depth) to obtain 3D points $\mathbf{X}_t^v=\Pi^{-1}(\mathbf{D}_t^v;\mathbf{K}^v,\mathbf{T}_t^v)$, and then predicts a set of Gaussians:
\begin{equation}
\mathcal{G}_t=\{(\boldsymbol{\mu}_i,\mathbf{a}_i)\}_{i=1}^{N}
\;\leftarrow\;
\Gamma_\psi(\mathbf{F}_t,\mathbf{X}_t),
\label{eq:existing_3dgs}
\end{equation}
where $\boldsymbol{\mu}_i$ denotes Gaussian centers and $\mathbf{a}_i$ summarizes the remaining Gaussian attributes for rendering, including opacity, scale, rotation, and appearance.
This makes the geometry proxy the interface between image features and renderable Gaussians.
Since the proxy geometry $\mathbf{X}_t$ provides the 3D support for Gaussian centers and visibility, it directly corresponds to the structure variable $\mathcal{X}_t$ in Eq.~\eqref{eq:task_render} up to parameterization.

This formulation exposes the limitation of head-centric designs.
Although Gaussian decoders can refine attributes and local parameters on top of a given support, they cannot fully correct globally inconsistent geometry when the upstream representation is weak.
The issue is therefore not the expressiveness of 3DGS itself, but the reliability of the support on which it is instantiated.
Under minimal overlap, locally plausible per-view depth may become inconsistent in the shared ego-centric space, and such errors are converted into misplaced Gaussians after lifting.
Thus, the main bottleneck lies before Gaussian prediction, as stable novel-view rendering requires input representations with stronger transferable geometry.

\subsection{Geometry foundation prior}
Our key hypothesis is that reliable inference of $\mathcal{X}_t$ in Eq.~\eqref{eq:task_render} should be supported by transferable upstream geometry rather than decoder side correction alone.
We therefore shift geometric reasoning to a geometry foundation backbone and adapt it to the surround-camera setting.
Specifically, we introduce a geometry backbone $\mathcal{B}_{\mathrm{geo}}$ to extract multi-view geometric features $\mathbf{H}_t=\mathcal{B}_{\mathrm{geo}}(\mathcal{I}_t)$, and a camera-conditioned adapter to predict both structure and appearance:
\begin{equation}
(\mathcal{X}_t,\mathcal{A}_t)
\;=\;
\Psi\!\big(\mathbf{H}_t;\{\mathbf{K}^v,\mathbf{T}_t^v\}_{v=1}^{V}\big).
\label{eq:foundation_adapt}
\end{equation}
Here $\mathbf{H}_t$ serves as an implicit structural prior for $\mathcal{X}_t$ under minimal overlap.
Rather than using geometric priors only as auxiliary depth cues, we treat them as representation-level priors that guide 3D structure inference before Gaussian decoding.
This is critical in sparse surround-view reconstruction, where explicit cross-view correspondences are scarce and local decoder constraints are insufficient.

However, directly transferring open-domain geometry priors to driving scenes is non-trivial.
Surround-view cameras have fixed but heterogeneous viewpoints, wide baselines, limited overlap, and meter-level ego-pose changes, which differ from the image configurations assumed by generic geometry models.
Thus, the pretrained prior must be adapted to preserve cross-view consistency, calibrate geometry to the ego-centric driving scale, and retain appearance cues for faithful rendering.
After adaptation, the inferred structure and appearance are converted into a renderable Gaussian scene:
\begin{equation}
\mathcal{G}_t
=
\Gamma_{\mathrm{gs}}\!\big(\mathcal{X}_t,\mathcal{A}_t\big),
\qquad
\hat I_\tau^v
=
\mathcal{R}\!\big(\mathcal{G}_t;\mathbf{K}^v,\mathbf{T}_\tau^v\big).
\label{eq:foundation_to_gs}
\end{equation}
Eq.~\eqref{eq:foundation_to_gs} makes explicit that geometry foundation priors affect rendering through the 3DGS support before novel-view synthesis, rather than only through decoder-side correction.
This establishes the role of geometry foundation priors and motivates the VGGD network instantiated in the following section.


\section{Network Design}
\label{sec:net}

\subsection{Overall}
\label{subsec:overall_design}
\begin{figure*}[t]
  \centering
  \includegraphics[width=1.0\linewidth]{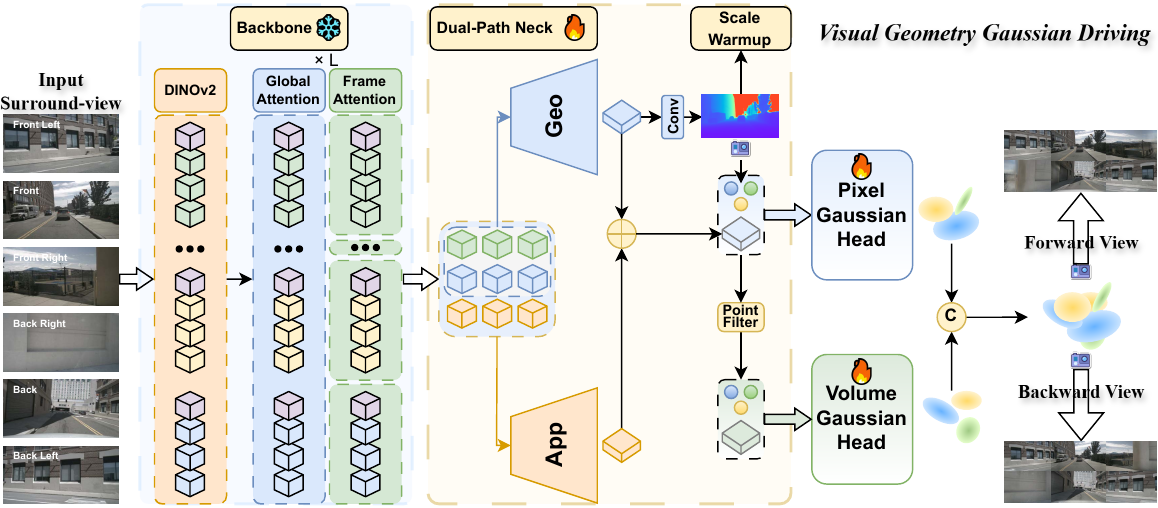}
  \caption{\textbf{Overview of VGGD.} Given surround images $\mathcal{I}_t=\{I_t^v\}$, the geometry foundation frontend $\Phi(\cdot)$ extracts multi-view priors $\mathbf{Z}_t=\Phi(\mathcal{I}_t)$. The driving-oriented adapter $\Psi(\cdot)$ predicts per-view depth $\mathbf{D}_t=\{\mathbf{D}_t^v\}$ and dense features $\mathbf{F}_t=\{\mathbf{F}_t^v\}$, and back-projects $\mathbf{D}_t$ to 3D points $\mathbf{X}_t=\{\mathbf{X}_t^v\}$, which instantiate the structure variable $\mathcal{X}_t$ in Eq.~\eqref{eq:task_render}. Finally, the Gaussian decoder $\Gamma(\cdot)$ outputs a renderable 3D Gaussian scene $\mathcal{G}_t=\Gamma(\mathbf{F}_t,\mathbf{X}_t)$, which is differentiably rendered to target views $\hat I_\tau^v$ at ego poses $\tau\in\mathcal{T}$.}
  \label{fig:arch}
\end{figure*}

As illustrated in Figure~\ref{fig:arch}, we propose \textbf{VGGD}, a \textbf{V}isual \textbf{G}eometry foundation-aware \textbf{G}aussian splatting framework for feed-forward surround-view \textbf{D}riving reconstruction.
Given the single-frame surround input $\mathcal{I}_t=\{I_t^v\}_{v=1}^{V}$ with camera parameters $\{\mathbf{K}^v,\mathbf{T}_t^v\}_{v=1}^{V}$, a geometry foundation frontend $\Phi(\cdot)$, which instantiates $\mathcal{B}_{\mathrm{geo}}$ in Eq.~\eqref{eq:foundation_adapt}, first extracts multi-view geometric priors $\mathbf{Z}_t=\{\mathbf{Z}_t^v\}_{v=1}^{V}$.
Here, $\mathbf{Z}_t$ corresponds to the geometry foundation representation $\mathbf{H}_t$ in Methodology.
A driving-oriented adaptation module $\Psi(\cdot)$ then converts these priors into the task-critical structure proxy and appearance features by predicting per-view depth $\mathbf{D}_t=\{\mathbf{D}_t^v\}_{v=1}^{V}$ and dense feature maps $\mathbf{F}_t=\{\mathbf{F}_t^v\}_{v=1}^{V}$.
The depth $\mathbf{D}_t^v$ is back-projected to 3D points $\mathbf{X}_t^v=\Pi^{-1}(\mathbf{D}_t^v;\mathbf{K}^v,\mathbf{T}_t^v)$, and $\mathbf{X}_t=\{\mathbf{X}_t^v\}_{v=1}^{V}$ serves as an explicit instantiation of the structure variable $\mathcal{X}_t$ in Eq.~\eqref{eq:task_render}.
Within $\Psi$, we employ a transient \emph{Scale Warmup} to calibrate the predicted geometry to the driving-camera configuration during early training, and a \emph{Dual-Path Neck} to produce geometry-consistent and appearance-enhanced features modeling $\mathcal{A}_t$.
Finally, a Gaussian decoder $\Gamma(\cdot)$ predicts a renderable 3D Gaussian scene $\mathcal{G}_t=\Gamma(\mathbf{F}_t,\mathbf{X}_t)$, where $\Gamma(\cdot)$ is the concrete 3DGS decoder corresponding to the adapted rendering formulation in Eq.~\eqref{eq:foundation_to_gs}.
The scene is differentiably rendered to target views $\hat I_\tau^v=\mathcal{R}(\mathcal{G}_t;\mathbf{K}^v,\mathbf{T}_\tau^v)$ for all $\tau\in\mathcal{T}$.


\subsection{VGGT Feature Extraction}
\label{subsec:vggt_feat}

We instantiate the geometry foundation frontend $\Phi(\cdot)$ with a pretrained VGGT model.
In the notation of Sec.~\ref{sec:method}, the extracted tokens $\mathbf{Z}_t$ provide the implementation of the upstream geometric representation $\mathbf{H}_t$.
Following VGGT feature tapping, we extract three types of per-view representations for each surround camera $v$:
(i) DINO-aligned features $\mathbf{Z}_{t}^{v,\mathrm{dino}}$ that preserve strong local descriptors,
(ii) frame features $\mathbf{Z}_{t}^{v,\mathrm{frm}}$ capturing view-specific context, and
(iii) global features $\mathbf{Z}_{t}^{v,\mathrm{glo}}$ encoding cross-view geometric context.
We group frame/global features as geometry-centric tokens
$\mathbf{Z}_{t}^{v,\mathrm{fg}}=[\mathbf{Z}_{t}^{v,\mathrm{frm}}\ \|\ \mathbf{Z}_{t}^{v,\mathrm{glo}}]$,
and form the full token used by the adapter as:
\[
\mathbf{Z}_{t}^{v}=\big[\mathbf{Z}_{t}^{v,\mathrm{fg}} \ \|\  \mathbf{Z}_{t}^{v,\mathrm{dino}}\big],
\qquad
\mathbf{Z}_t=\{\mathbf{Z}_t^v\}_{v=1}^{V}=\Phi(\mathcal{I}_t).
\]
We additionally retain $\mathbf{Z}_{t}^{v,\mathrm{dino}}$ because the aggregator layer of VGGT is primarily geometry-driven and tends to compress semantic and high-frequency appearance cues, which are crucial for modeling $\mathcal{A}_t$ in sparse surround-view reconstruction.
We freeze $\Phi$ during training and optimize only the downstream adaptation and Gaussian decoding modules.

\subsection{Geometry and Appearance Modeling}
\label{subsec:geo_app_learning}

\subsubsection{Dual-Path Design}
To instantiate Eq.~\eqref{eq:task_render}, the adapter $\Psi(\cdot)$ must infer both the structure $\mathcal{X}_t$ and the appearance $\mathcal{A}_t$ from a single sparse surround capture.
In extremely low-overlap settings, geometry-centric features provide more stable 3D support but may suppress semantic and high-frequency details, while appearance-rich features alone lack reliable geometric grounding.
We therefore introduce a \emph{Dual-Path Neck} with parallel geometry and appearance paths inside $\Psi(\cdot)$:
\begin{equation}
(\mathbf{D}_t^v,\mathbf{F}_{t}^{v,\mathrm{geo}})=\Psi_{\mathrm{geo}}\!\left(\mathbf{Z}_{t}^{v,\mathrm{fg}}\right),
\qquad
\mathbf{F}_{t}^{v,\mathrm{app}}=\Psi_{\mathrm{app}}\!\left(\mathbf{Z}_{t}^{v}\right),
\label{eq:dual_path}
\end{equation}
where $\mathbf{D}_t^v$ provides the structure proxy and $\mathbf{F}_{t}^{v,\mathrm{geo}}$/$\mathbf{F}_{t}^{v,\mathrm{app}}$ denote geometry-consistent and appearance-aware features, respectively.
We fuse the two paths by element-wise addition, i.e., $\mathbf{F}_t^v=\mathbf{F}_{t}^{v,\mathrm{geo}}+\mathbf{F}_{t}^{v,\mathrm{app}}$, and use $\mathbf{F}_t=\{\mathbf{F}_t^v\}_{v=1}^{V}$ as the feature input to subsequent Gaussian heads.
This design keeps $\mathbf{F}_{t}^{v,\mathrm{geo}}$ aligned with the inferred structure, while $\mathbf{F}_{t}^{v,\mathrm{app}}$ preserves semantic and high-frequency cues for modeling $\mathcal{A}_t$.

We back-project the predicted depth to 3D points $\mathbf{X}_t^v=\Pi^{-1}(\mathbf{D}_t^v;\mathbf{K}^v,\mathbf{T}_t^v)$, yielding $\mathbf{X}_t=\{\mathbf{X}_t^v\}_{v=1}^{V}$, which explicitly instantiates the structure variable $\mathcal{X}_t$ up to parameterization.
Following Omni-Scene-style bounded completion, we define a driving-centric cuboid range $\mathcal{B}\subset\mathbb{R}^3$ and construct an in-range mask $\mathbf{M}_t^v(\mathbf{u})=\mathbb{I}[\mathbf{X}_t^v(\mathbf{u})\in\mathcal{B}]$ to select valid points and features for the volume branch:
\begin{equation}
\mathbf{X}_{t,\mathrm{vol}}^v=\mathbf{X}_t^v\odot\mathbf{M}_t^v,
\qquad
\mathbf{F}_{t,\mathrm{vol}}^v=\mathbf{F}_t^v\odot\mathbf{M}_t^v .
\label{eq:volume_mask}
\end{equation}
Specifically, the pixel branch uses the full $(\mathbf{X}_t^v,\mathbf{F}_t^v)$ to preserve image-level details, while the volume branch focuses on $(\mathbf{X}_{t,\mathrm{vol}}^v,\mathbf{F}_{t,\mathrm{vol}}^v)$ for bounded 3D completion.

\subsubsection{Scale Warmup Strategy.}
The task is highly sensitive to the structure variable $\mathcal{X}_t$ in Eq.~\eqref{eq:task_render}, since small scale drift in predicted depth can be amplified under meter-level ego-pose changes and lead to misalignment in novel views.
Although geometry foundation priors provide strong structural cues, their open-domain pretraining does not necessarily match the driving camera configuration or metric scale.
Our motivation is to mitigate the initial scale ambiguity: early anchoring stabilizes the back-projected geometry, while releasing the constraint enables end-to-end refinement beyond the coarse reference.

Specifically, we apply \emph{Scale Warmup} as transient supervision on the geometry-path depth prediction $\mathbf{D}_t^v$.
Let $\tilde{\mathbf{D}}_{t}^{v}$ denote a coarse depth reference for $I_t^v$.
During the first $S_{\mathrm{warm}}$ steps, we use a gated $\mathcal{L}_1$ loss:
\begin{equation}
\mathcal{L}_{\mathrm{warm}}(s)
=
\mathbb{I}[s \le S_{\mathrm{warm}}]\;
\sum_{v=1}^{V}
\left\|\mathbf{D}_{t}^{v}-\tilde{\mathbf{D}}_{t}^{v}\right\|_{1}.
\label{eq:warmup_loss}
\end{equation}
This early anchoring calibrates the scale of the back-projected points $\mathbf{X}_t$ and improves cross-view stability of $\mathcal{X}_t$, providing a reliable geometric substrate for subsequent 3DGS decoding.
After warmup, the direct alignment on the geometry-path input depth is disabled, while the rendered-depth losses remain as output-space geometric regularizers.
This avoids persistent constraints on $\mathbf{D}_t^v$ while retaining task-level depth supervision for end-to-end refinement. The complete training procedure, including the transient warmup schedule and the full rendering objective, is summarized in Algorithm~\ref{alg:scale_warmup}.


\subsection{Gaussian Parameter Decoder}
\label{subsec:gaussian_decoder}

We parameterize the latent scene $S_t$ in Eq.~\eqref{eq:task_render} with a hybrid pixel--volume 3DGS representation, following Omni-Scene.
This decoder instantiates $\Gamma(\cdot)$ in Sec.~\ref{subsec:overall_design} with two complementary Gaussian heads.
This decomposition separates image-aligned detail preservation from bounded-space completion, matching the weak-overlap driving setting.
It takes the adapted outputs of $\Psi(\cdot)$ from Sec.~\ref{subsec:geo_app_learning}: the fused feature maps $\mathbf{F}_t=\{\mathbf{F}_t^v\}_{v=1}^{V}$, the structure proxy $\mathbf{X}_t=\{\mathbf{X}_t^v\}_{v=1}^{V}$, and the in-range mask $\mathbf{M}_t=\{\mathbf{M}_t^v\}_{v=1}^{V}$ defined by the point-cloud range $\mathcal{B}$, and converts them into a renderable Gaussian set $\mathcal{G}_t$.

Different from Omni-Scene, which filters pixel-decoded features before feeding the volume branch, we route neck features directly to both heads.
Specifically, the pixel head consumes the full $(\mathbf{F}_t,\mathbf{X}_t)$ to preserve image-space details and far-range content, while the volume head consumes only the masked subsets $(\mathbf{F}_t\odot\mathbf{M}_t,\ \mathbf{X}_t\odot\mathbf{M}_t)$ to focus on bounded 3D completion within $\mathcal{B}$.
This removes an explicit pixel-to-volume feature dependency and allows both heads to benefit from our geometry-aligned and appearance-enhanced neck outputs.
We use these two Gaussian heads, $\Gamma_{\mathrm{pix}}$ and $\Gamma_{\mathrm{vol}}$, to predict Gaussians in pixel space and bounded volume space, respectively.
The final Gaussian scene is obtained by:
\begin{equation}
\mathcal{G}_t
\;=\;
\Gamma_{\mathrm{pix}}(\mathbf{F}_t,\mathbf{X}_t)
\ \cup\
\Gamma_{\mathrm{vol}}(\mathbf{F}_t\odot \mathbf{M}_t,\mathbf{X}_t\odot \mathbf{M}_t).
\label{eq:final_gauss}
\end{equation}
where $\cup$ denotes concatenation of the pixel and volume Gaussian sets.
In practice, the pixel head captures fine details and far-range content, while the volume head improves completeness within the driving-centric range.


\begin{algorithm}[!t]
\caption{Training with transient Scale Warmup}
\label{alg:scale_warmup}
\begin{algorithmic}[1]
\REQUIRE Surround images $\mathcal{I}_t=\{I_t^v\}_{v=1}^{V}$, camera parameters $\{\mathbf{K}^v,\mathbf{T}_t^v\}_{v=1}^{V}$, target views $\{I_\tau^v,\mathbf{T}_\tau^v \mid \tau\in\mathcal{T},v=1,\ldots,V\}$, coarse depth references $\{\tilde{\mathbf{D}}_t^v,\tilde{\mathbf{D}}_\tau^v\}_{v=1}^{V}$, warmup steps $S_{\mathrm{warm}}$
\FOR{training step $s=1,2,\ldots$}
    \STATE $\mathbf{Z}_t \leftarrow \Phi(\mathcal{I}_t)$
    \STATE $(\mathbf{D}_t,\mathbf{F}_t) \leftarrow \Psi(\mathbf{Z}_t;\{\mathbf{K}^v,\mathbf{T}_t^v\}_{v=1}^{V})$
    \STATE $\mathbf{X}_t \leftarrow \{\Pi^{-1}(\mathbf{D}_t^v;\mathbf{K}^v,\mathbf{T}_t^v)\}_{v=1}^{V}$
    \STATE $\mathbf{M}_t \leftarrow \{\mathbf{M}_t^v\}_{v=1}^{V},\quad \mathbf{M}_t^v(\mathbf{u})=\mathbb{I}[\mathbf{X}_t^v(\mathbf{u})\in\mathcal{B}]$
    \STATE $\mathcal{G}_{t}^{\mathrm{pix}} \leftarrow \Gamma_{\mathrm{pix}}(\mathbf{F}_t,\mathbf{X}_t)$
    \STATE $\mathcal{G}_{t}^{\mathrm{vol}} \leftarrow \Gamma_{\mathrm{vol}}(\mathbf{F}_t\odot\mathbf{M}_t,\mathbf{X}_t\odot\mathbf{M}_t)$
    \STATE $\mathcal{G}_t \leftarrow \mathcal{G}_{t}^{\mathrm{pix}}\cup\mathcal{G}_{t}^{\mathrm{vol}}$
    \STATE $\hat I_\tau^v \leftarrow \mathcal{R}(\mathcal{G}_t;\mathbf{K}^v,\mathbf{T}_\tau^v), \quad \forall \tau,v$
    \STATE $\hat{\mathbf{D}}_\tau^v \leftarrow \mathcal{R}_{\mathrm{dep}}(\mathcal{G}_t;\mathbf{K}^v,\mathbf{T}_\tau^v), \quad \forall \tau,v$
    \STATE Compute $\mathcal{L}_{\mathrm{rgb}}$, $\mathcal{L}_{\mathrm{perc}}$, $\mathcal{L}_{\mathrm{dep}}$, $\mathcal{L}_{\mathrm{rgb}}^{\mathrm{vol}}$, and $\mathcal{L}_{\mathrm{dep}}^{\mathrm{vol}}$
    \STATE $\mathcal{L}_{\mathrm{warm}}(s) \leftarrow
    \mathbb{I}[s \le S_{\mathrm{warm}}]\sum_{v=1}^{V}
    \left\|\mathbf{D}_{t}^{v}-\tilde{\mathbf{D}}_{t}^{v}\right\|_{1}$
    \STATE $\mathcal{L} \leftarrow
    \mathcal{L}_{\mathrm{rgb}}
    + \lambda_{\mathrm{perc}}\mathcal{L}_{\mathrm{perc}}
    + \lambda_{\mathrm{dep}}\mathcal{L}_{\mathrm{dep}}
    + \mathcal{L}_{\mathrm{rgb}}^{\mathrm{vol}}
    + \lambda_{\mathrm{dep}}^{\mathrm{vol}}\mathcal{L}_{\mathrm{dep}}^{\mathrm{vol}}
    + \lambda_{\mathrm{warm}}\mathcal{L}_{\mathrm{warm}}(s)$
    \STATE Update trainable parameters of $\Psi$, $\Gamma_{\mathrm{pix}}$, and $\Gamma_{\mathrm{vol}}$ by minimizing $\mathcal{L}$
\ENDFOR
\end{algorithmic}
\end{algorithm}

\subsection{Training Strategy}
\label{subsec:training_strategy}

Given the predicted Gaussian scene $\mathcal{G}_t$ from Eq.~\eqref{eq:final_gauss}, we render target views at displaced ego poses $\tau\in\mathcal{T}$ via
$\hat I_\tau^v=\mathcal{R}(\mathcal{G}_t;\mathbf{K}^v,\mathbf{T}_\tau^v)$.
We supervise novel-view rendering with a photometric $\mathcal{L}_1$ loss $\mathcal{L}_{\mathrm{rgb}}$ and a perceptual LPIPS loss $\mathcal{L}_{\mathrm{perc}}$.
We also render depth maps $\hat{\mathbf{D}}_\tau^v=\mathcal{R}_{\mathrm{dep}}(\mathcal{G}_t;\mathbf{K}^v,\mathbf{T}_\tau^v)$ and apply a depth loss $\mathcal{L}_{\mathrm{dep}}$ against a coarse depth reference $\tilde{\mathbf{D}}_\tau^v$.
These losses supervise both appearance fidelity and geometric consistency across target cameras and displaced poses.

To strengthen bounded-range completion, we render only the volume Gaussian set $\mathcal{G}_{t}^{\mathrm{vol}}=\Gamma_{\mathrm{vol}}(\mathbf{F}_t\odot \mathbf{M}_t,\mathbf{X}_t\odot \mathbf{M}_t)$ and compute masked $\mathcal{L}_1$ and depth losses within the point-cloud range $\mathcal{B}$, denoted as $\mathcal{L}_{\mathrm{rgb}}^{\mathrm{vol}}$ and $\mathcal{L}_{\mathrm{dep}}^{\mathrm{vol}}$.
\emph{Scale Warmup} further adds an early-stage depth alignment term $\mathcal{L}_{\mathrm{warm}}(s)$ on $\mathbf{D}_t^v$ in Eq.~\eqref{eq:warmup_loss} for the first $S_{\mathrm{warm}}$ training steps.
The overall training objective is:
\begin{equation}
\begin{split}
\mathcal{L}
= \, &\mathcal{L}_{\mathrm{rgb}}
+ \lambda_{\mathrm{perc}}\mathcal{L}_{\mathrm{perc}}
+ \lambda_{\mathrm{dep}}\mathcal{L}_{\mathrm{dep}} \\
+ \, &\mathcal{L}_{\mathrm{rgb}}^{\mathrm{vol}}
+ \lambda_{\mathrm{dep}}^{\mathrm{vol}}\mathcal{L}_{\mathrm{dep}}^{\mathrm{vol}}
+ \lambda_{\mathrm{warm}}\mathcal{L}_{\mathrm{warm}}(s).
\end{split}
\label{eq:total_loss}
\end{equation}
We set the loss weights $\lambda_{\mathrm{perc}}$, $\lambda_{\mathrm{dep}}$, $\lambda_{\mathrm{dep}}^{\mathrm{vol}}$, and $\lambda_{\mathrm{warm}}$ to $0.05,0.01,0.01,0.01$, respectively.

\section{Experiments}
\label{sec:exp}

\subsection{Experimental Setup}
\label{subsec:exp_setup}

\input{tables/quan}

\subsubsection{Datasets}
We follow the single-frame surround-view reconstruction protocol of Omni-Scene~\cite{wei2025omniscene} on nuScenes~\cite{caesar2020nuscenes}, which contains 700 training and 150 validation scenes recorded at 12\,Hz for approximately 20 seconds.
The center frame provides $V{=}6$ surround-view images as inputs, while the two endpoint frames provide $12$ novel target views, with $6$ center views for supervision and evaluation.
This yields 135,941 training bins and 30,080 validation bins.
For a fair comparison, all methods are evaluated at the same resolution.

\subsubsection{Evaluation Metrics}
We report standard photometric and perceptual metrics for novel-view synthesis, including PSNR, SSIM~\cite{wang2004ssim}, and LPIPS~\cite{zhang2018lpips}.
To assess 3D structural consistency, we render depth maps and compute the Pearson Correlation Coefficient (PCC)~\cite{benesty2009pearson} with respect to pseudo relative depth maps predicted by Depth Anything V2~\cite{yang2024depthanythingv2}, as in Omni-Scene~\cite{wei2025omniscene}.
Since PCC measures linear correlation, it is invariant to global shift and positive scaling, making it suitable for evaluating relative depth consistency under ego-pose changes.
Unless otherwise stated, all metrics are computed over the 18 rendered views per sample and averaged over the entire validation set.

\subsubsection{Baseline Methods}
\label{subsec:baselines}
We compare against baselines from two paradigms: per-scene optimization and generalizable feed-forward reconstruction.
We report quantitative results following the protocols in~\cite{wei2025omniscene,lu2025phigenesis}.
For optimization-based reconstruction methods, we include EmerNeRF~\cite{yang2024emernerf} and 3DGS~\cite{kerbl20233dgs}, as well as dynamic Gaussian fitting variants PVG~\cite{chen2023pvg} and DeformableGS~\cite{yang2024deformable3dgs}, which optimize scene-specific parameters and serve as strong reference points under the same evaluation protocol.
For feed-forward baselines, we include generalizable neural rendering methods, including AttnRend~\cite{Du_2023_CVPR} and MuRF~\cite{Xu_2024_CVPR}, as well as Gaussian regression/reconstruction models, including pixelSplat~\cite{charatan2024pixelsplat} and MVSplat~\cite{chen2024mvsplat}.
We also compare with driving-oriented methods, including DrivingForward~\cite{tian2024drivingforward} and Omni-Scene~\cite{wei2025omniscene}.
We further include large-scale feed-forward paradigms that emphasize scalability via Transformer or diffusion backbones, including LGM~\cite{tang2024lgm}, GS-LRM~\cite{zhang2024gslrm}, SCube~\cite{ren2024scube}, and STORM~\cite{yang2025storm}.
All baselines are evaluated under the same single-frame rendering protocol.

\subsubsection{Implementation Details}
We train the network using AdamW with a cosine annealing schedule, decaying the learning rate from $1{\times}10^{-4}$ to $1{\times}10^{-6}$.
The model is trained for 100,000 iterations.
Following Omni-Scene~\cite{wei2025omniscene}, we supervise rendered depth using coarse metric depth pseudo labels predicted by Metric3D-v2~\cite{hu2024metric3dv2} through the depth loss term in our final objective.
We apply \emph{Scale Warmup} to geometry-path depth for the first 2,000 iterations and implement the framework in PyTorch with bfloat16.
All other unstated hyperparameters align with the Omni-Scene~\cite{wei2025omniscene} configuration.

\begin{figure*}[t!]
  \centering
  \includegraphics[width=0.985\linewidth]{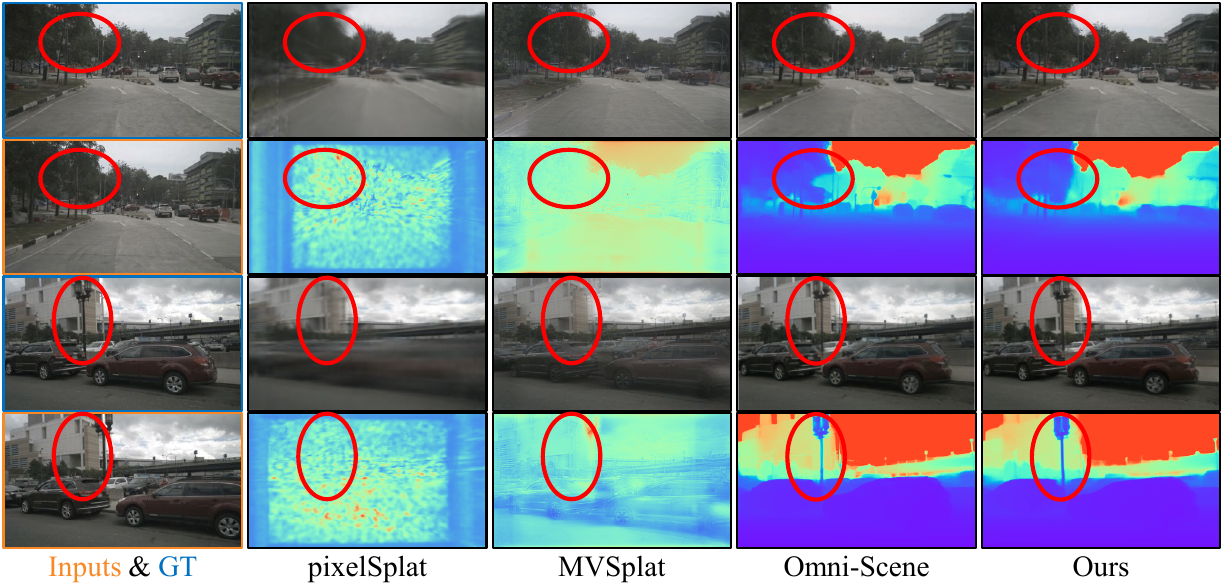} 
  \caption{Qualitative comparisons of single-view details. Input views are highlighted with orange boxes, and ground-truth target views are highlighted with blue boxes. All other panels show the synthesized novel views and the corresponding depth maps, where warmer colors indicate larger distances and cooler colors indicate smaller distances.}
  \label{fig:exp1}
\end{figure*}

\subsection{Quantitative Comparisons}
\label{subsec:quantitative_comparisons}

\subsubsection{Per-scene optimization methods}
As shown in Table~\ref{tab:main_results}, VGGD substantially outperforms per-scene optimization methods, including EmerNeRF~\cite{yang2024emernerf}, 3DGS~\cite{kerbl20233dgs}, PVG~\cite{chen2023pvg}, and DeformableGS~\cite{yang2024deformable3dgs}.
Although these methods optimize scene-specific parameters, they remain brittle under sparse surround-view observations and ego-pose changes.
Compared with DeformableGS, the strongest optimization-based baseline in the table, VGGD improves PSNR by 4.73 dB and PCC by 0.703.
This large gap highlights the difficulty of fitting consistent 3D structure from weakly overlapping views alone.
These results indicate that per-scene fitting is less effective under the minimal-overlap protocol, whereas the complete VGGD framework benefits from transferable upstream representations.

\subsubsection{General-purpose feed-forward methods}
VGGD also consistently outperforms general-purpose feed-forward baselines.
AttnRend~\cite{Du_2023_CVPR} and MuRF~\cite{Xu_2024_CVPR} remain limited in rendering quality, with PSNR scores of 20.96 and 20.34, respectively, and MuRF further shows unreliable geometry with negative PCC.
Gaussian regression methods, including pixelSplat~\cite{charatan2024pixelsplat} and MVSplat~\cite{chen2024mvsplat}, improve efficiency but still perform poorly in the weak-overlap driving setting, reaching only 21.51/21.61 PSNR and up to 0.181 PCC.
Large-scale models such as LGM~\cite{tang2024lgm}, GS-LRM~\cite{zhang2024gslrm}, and SCube~\cite{ren2024scube} improve over smaller generalizable baselines, but their best results remain below VGGD across all metrics.
The results suggest that the complete driving-oriented VGGD design is better suited to the minimal-overlap setting than generic sparse-view transfer or capacity scaling.

\subsubsection{Driving-centric pipelines}
We further compare against driving-centric pipelines under the target sparse single-frame surround-view setting.
Among driving-centric methods, VGGD achieves the best overall performance on the single-frame surround-view reconstruction benchmark.
It improves over DrivingForward~\cite{tian2024drivingforward} by +0.53 PSNR (24.32$\rightarrow$\textbf{24.85}), +0.026 SSIM (0.732$\rightarrow$\textbf{0.758}), $\downarrow$0.013 LPIPS (0.229$\rightarrow$\textbf{0.216}), and +0.042 PCC (0.766$\rightarrow$\textbf{0.808}).
Compared with Omni-Scene~\cite{wei2025omniscene}, VGGD gains +0.58 PSNR, +0.022 SSIM, and $\downarrow$0.021 LPIPS, while slightly improving PCC from 0.804 to \textbf{0.808}.
VGGD also surpasses STORM~\cite{yang2025storm}, a large spatio-temporal reconstruction model, by +0.29 PSNR and +0.020 PCC.
Although STORM uses spatio-temporal modeling, the reported VGGD results remain higher in PSNR and PCC using only a single surround frame.
This comparison highlights the strength of single-frame geometric representation learning.
Together, these results validate VGGD's driving-oriented geometry foundation adaptation for stable, high-fidelity, and geometrically consistent reconstruction under low-overlap single-frame driving constraints.

\begin{figure*}[t!]
  \centering
  \includegraphics[width=0.985\linewidth]{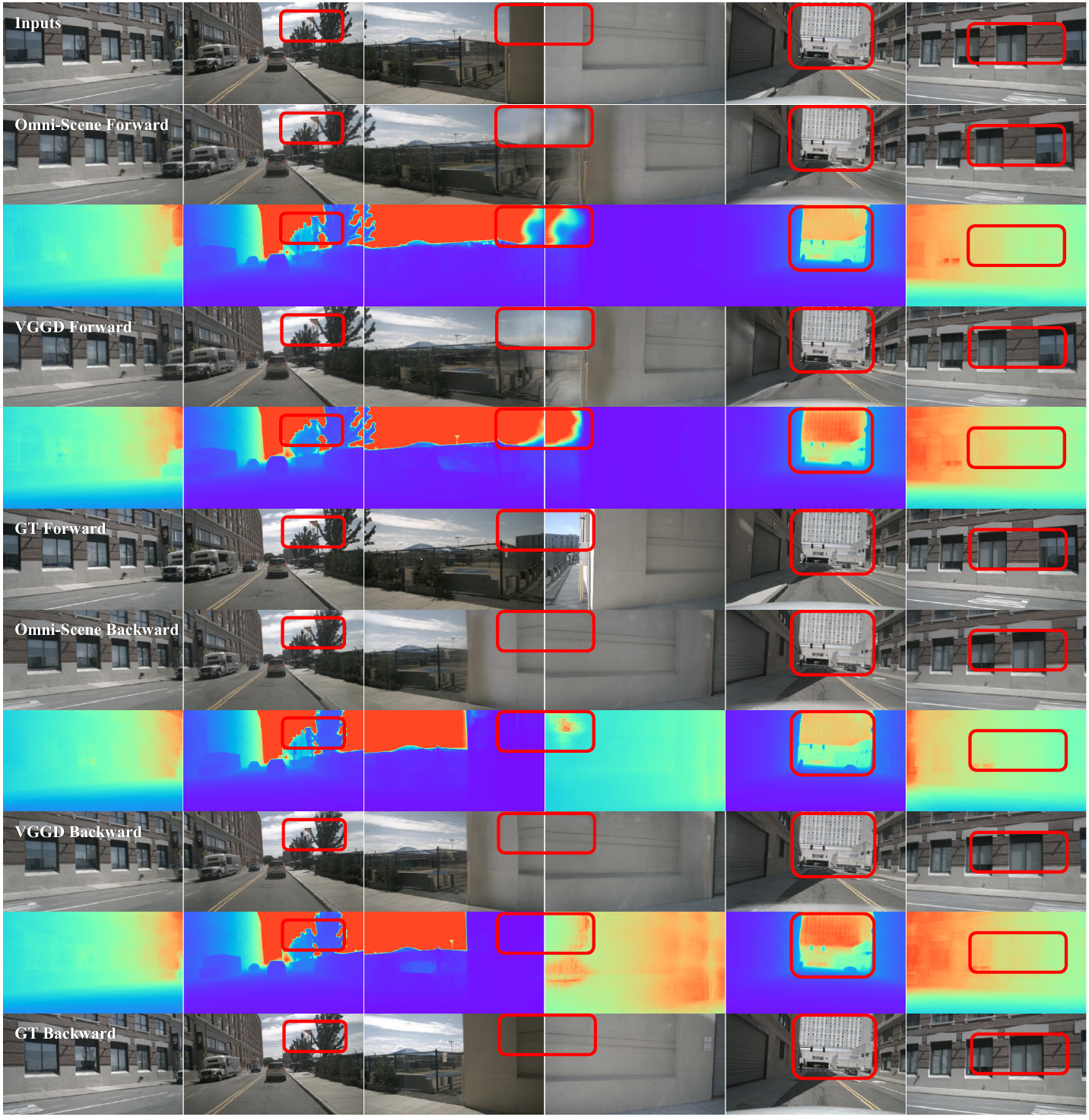} 
  \caption{Qualitative Comparisons of surround-views. A comparison of the input views and the corresponding forward/backward novel-view renderings and depth maps is shown. Red rounded rectangles highlight the regions with noticeable differences.}
  \label{fig:exp2}
\end{figure*}


\subsection{Qualitative Comparisons}
\label{subsec:qualitative}

\subsubsection{Single-view details}
We assess single-view fidelity by examining whether novel views preserve fine structures and plausible textures in weakly observed regions.
Figure~\ref{fig:exp1} presents a representative target view together with the corresponding depth maps to highlight local failure modes under extreme surround-view sparsity.
pixelSplat and MVSplat often produce blurred textures and collapsed structures, and their depth maps exhibit noisy patterns or over-smoothed boundaries around thin objects and distant backgrounds.
Omni-Scene improves overall realism, but still shows localized artifacts in the cropped regions, including boundary bleeding and depth irregularities that correlate with visible distortions.
In contrast, VGGD preserves sharper edges and higher-frequency details, while maintaining cleaner depth boundaries and a more coherent scene layout.
These observations suggest that VGGD better couples fine-grained appearance recovery with geometry-consistent rendering under weak overlap.

\subsubsection{Full surround-view comparison}
We evaluate cross-view geometric consistency by checking whether renderings and rendered depth maps remain stable across all cameras under meter-level ego-pose changes.
Figure~\ref{fig:exp2} compares VGGD with Omni-Scene on full surround-view renderings under forward and backward ego-pose changes, along with the corresponding rendered depth for geometric consistency.
The highlighted regions show that Omni-Scene can suffer from view-inconsistent artifacts and depth discontinuities across cameras, which become more pronounced at boundary views with larger viewpoint displacement.
VGGD yields more stable renderings across the six cameras and produces depth maps with smoother, more consistent transitions that better match the ground truth in both directions, especially for distant structures and low-overlap areas.
These observations indicate that VGGD produces more robust single-frame surround-view renderings under the evaluated ego-pose changes.

\begin{figure*}[t]
  \centering
  \includegraphics[width=1\linewidth]{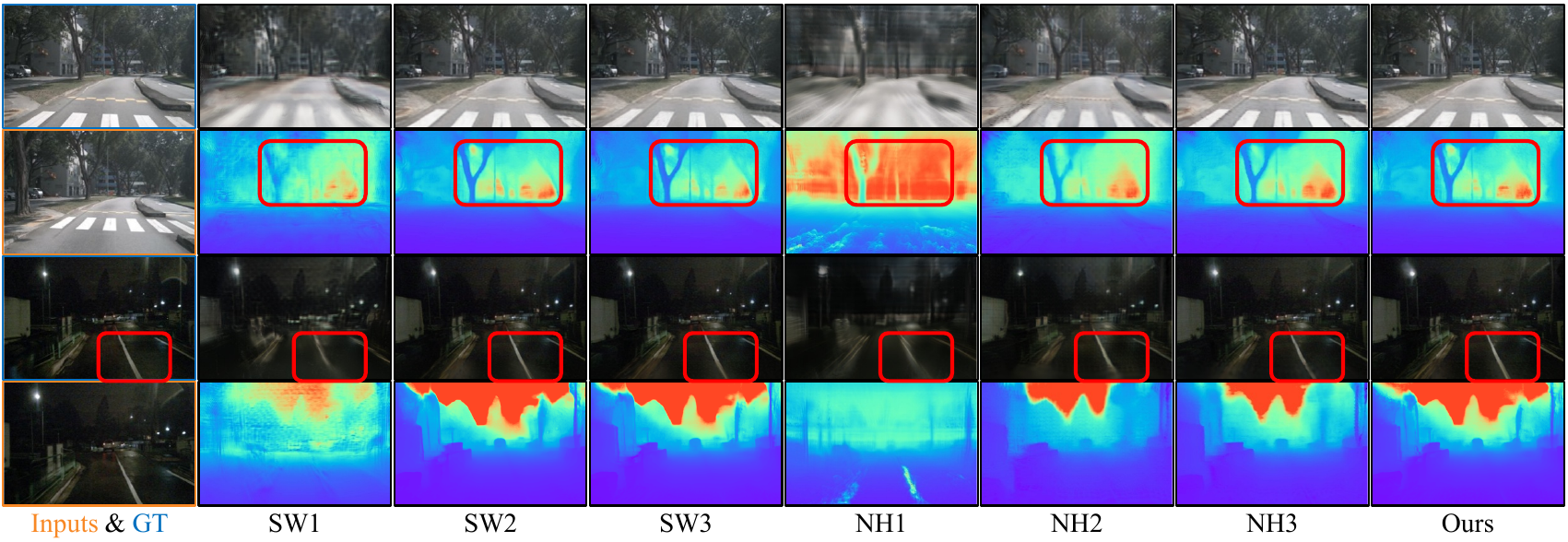} 
  \caption{Qualitative comparisons of ablation studies. Input views are highlighted with orange boxes, and ground-truth target views are highlighted with blue boxes. All other panels show the synthesized novel views and the corresponding depth maps, where warmer colors indicate larger distances and cooler colors indicate smaller distances.}
  \label{fig:exp3}
\end{figure*}

\input{tables/abla}


\subsection{Ablation Studies}
\label{subsec:ablation}

\subsubsection{Ablation setup}
To ensure a controlled ablation study, we train and evaluate all variants on the same nuScenes split, with identical input resolution, data preprocessing, and initialization.
Each run is trained for 10k iterations using AdamW and a cosine annealing learning-rate schedule decaying from $1{\times}10^{-4}$ to $1{\times}10^{-5}$.
All losses, augmentations, and remaining hyperparameters are kept fixed across variants.

\subsubsection{Scale Warmup Ablation}
We ablate the scale alignment schedule using three variants and compare them with our full model: \textbf{SW1-None} removes warmup, \textbf{SW2-Always} keeps warmup throughout training, \textbf{SW3-Half} applies warmup for the first half of training iterations, and \textbf{Full VGGD} uses transient warmup for the first $S_{\mathrm{warm}}{=}2000$ iterations.
As shown in Table~\ref{tab:abla}, removing warmup leads to a severe drop in both rendering and geometry (21.92 PSNR, 0.454 LPIPS, 0.755 PCC), indicating unstable scale under meter-level ego-pose changes.
All warmup-enabled variants substantially recover performance, bringing PSNR to around 24.0 and PCC to around 0.814.
Notably, persistent warmup provides no further gain in geometry and slightly compromises appearance, highlighting the benefit of a transient schedule.
Our \textbf{Full VGGD} achieves the best appearance quality with 24.12 PSNR and 0.247 LPIPS, while matching the best geometric consistency with 0.814 PCC.
This trend is also reflected in the corresponding panels of Figure~\ref{fig:exp3}, where removing warmup yields warped structures and noisy depth, whereas transient warmup produces cleaner boundaries and more stable depth.
Overall, the ablation validates \emph{Scale Warmup} as an effective component in VGGD.

\subsubsection{Neck\&Head Ablation}
We analyze feature routing and Gaussian decoding with three variants.
Here, direct GS predicts Gaussian parameters using a single convolution layer followed by the corresponding activations.
\textbf{NH1-Geo-PV} removes the \emph{Dual-Path Neck} and feeds only geometry features into the pixel--volume decoder, resulting in the worst performance (20.47 PSNR, 0.527 LPIPS, 0.619 PCC) and blurry renderings with noisy depth in Figure~\ref{fig:exp3}.
\textbf{NH2-Geo-GS} applies direct GS to geometry features, improving PCC to 0.764 but still yielding poor appearance quality with 0.371 LPIPS, consistent with over-smoothed textures.
\textbf{NH3-Geo\&App-GS} incorporates both geometry and appearance features into direct GS, which improves rendering quality (23.83 PSNR, 0.271 LPIPS) but degrades geometry (0.681 PCC), producing less coherent depth maps.
In contrast, \textbf{Full VGGD} achieves the best overall trade-off (24.12 PSNR, 0.247 LPIPS, 0.814 PCC), validating the importance of combining the \emph{Dual-Path Neck} with pixel--volume decoding.
Overall, these ablations show that the neck, as the bridge between the backbone and the GS head, has a substantial impact on final quality, supporting our central claim that proper feature decoupling and routing are crucial for geometry-consistent reconstruction.


\section{Conclusion and Future Works}
\label{sec:conclusion}

We propose \textbf{VGGD}, a visual geometry foundation-aware 3DGS framework for single-frame surround-view driving reconstruction. 
VGGD leverages pretrained geometric priors to mitigate the severe ambiguity caused by minimal camera overlap, and introduces driving-oriented architecture to bridge the domain gap.
To address geometry and appearance modeling, we introduce \emph{Scale Warmup} to stabilize metric scale learning and a \emph{Dual-Path Neck} to bridge the backbone and Gaussian decoder by decoupling geometry-consistent and appearance-aware representations. 
With a standard hybrid pixel--volume 3DGS decoder, we achieve state-of-the-art novel-view synthesis quality and improved geometric consistency on the nuScenes single-frame benchmark, supported by extensive quantitative, qualitative, and ablation studies.

Our core insight is that single-frame sparse surround-view reconstruction benefits from geometry-aware upstream representations rather than decoder-centric design alone.
We hope VGGD motivates further exploration of geometry-aware foundation priors for robust and scalable feed-forward reconstruction in real-world driving scenes.
A current limitation is that our framework does not model scene dynamics, leaving dynamic-aware reconstruction as an important future direction.



{
\small
\bibliographystyle{IEEEtran}
\bibliography{refs}
}

\end{document}

%% file: tables/quan.tex
\begin{table*}[!t]
\centering
\caption{\textbf{Quantitative results on nuScenes single-frame ego-centric surround-view reconstruction.}
We report PSNR, SSIM, and LPIPS for reconstruction and novel-view rendering quality, together with PCC for relative geometric consistency. (higher is better for PSNR/SSIM/PCC, lower is better for LPIPS).
STORM is included as a spatio-temporal reference.
Best and second-best results are highlighted in \textbf{bold} and \underline{underline}, respectively.}
\label{tab:main_results}
\resizebox{0.85\textwidth}{!}{
\begin{tabular}{c|c|c|cccc}
\toprule
\textbf{Category} & \textbf{Method} & \textbf{Pub\&Year} & \textbf{PSNR} $\uparrow$ & \textbf{SSIM} $\uparrow$ & \textbf{LPIPS} $\downarrow$ & \textbf{PCC} $\uparrow$ \\
\midrule
\multirow{4}{*}{\begin{tabular}[c]{@{}l@{}}\textit{Per-Scene}\\ \textit{Optimization}\end{tabular}}
& EmerNeRF~\cite{yang2024emernerf} & ICLR'24 & 18.45 & 0.582 & 0.502 & 0.061 \\
& 3DGS~\cite{kerbl20233dgs} & SIGGRAPH'23 & 19.67 & 0.603 & 0.436 & 0.094 \\
& PVG~\cite{chen2023pvg} & IJCV'26 & 18.98 & 0.567 & 0.481 & 0.072 \\
& DeformableGS~\cite{yang2024deformable3dgs} & CVPR'24 & 20.12 & 0.622 & 0.422 & 0.105 \\
\midrule
\multirow{11}{*}{\cellcolor{white}\begin{tabular}[c]{@{}l@{}}\textit{Generalizable}\\ \textit{Feed-Forward}\end{tabular}}
& AttnRend~\cite{Du_2023_CVPR} & CVPR'23 & 20.96 & 0.533 & 0.467 & N/A \\
& MuRF~\cite{Xu_2024_CVPR} & CVPR'24 & 20.34 & 0.504 & 0.433 & -0.332 \\
& LGM~\cite{tang2024lgm} & ECCV'24 & 22.15 & 0.672 & 0.318 & 0.342 \\
& GS-LRM~\cite{zhang2024gslrm} & ECCV'24 & 23.41 & 0.703 & 0.273 & 0.598 \\
& pixelSplat~\cite{charatan2024pixelsplat} & CVPR'24 & 21.51 & 0.616 & 0.372 & 0.001 \\
& MVSplat~\cite{chen2024mvsplat} & ECCV'24 & 21.61 & 0.658 & 0.295 & 0.181 \\
& SCube~\cite{ren2024scube} & NeurIPS'24 & 23.85 & 0.721 & 0.258 & 0.651 \\
& DrivingForward~\cite{tian2024drivingforward} & AAAI'25 & 24.32 & 0.732 & 0.229 & 0.766 \\
& Omni-Scene~\cite{wei2025omniscene} & CVPR'25 & 24.27 & 0.736 & 0.237 & \underline{0.804} \\
& STORM~\cite{yang2025storm} & ICLR'25 & \underline{24.56} & \underline{0.752} & \underline{0.217} & 0.788 \\
\cmidrule{2-7}
\rowcolor{gray!10} \cellcolor{white} & \textbf{VGGD} & Ours &\textbf{24.85} & \textbf{0.758} & \textbf{0.216} & \textbf{0.808}  \\
\bottomrule
\end{tabular}
}
\end{table*}

%% file: tables/abla.tex
\begin{table}[t!]
\centering

\caption{\textbf{Ablation study on nuScenes.} 
We evaluate (i) \emph{Scale Warmup} schedules (SW1--SW3) and (ii) neck\&head designs (NH1--NH3).
Our full model uses \emph{Scale Warmup} with $S_{\mathrm{warm}}{=}2000$ iterations and the proposed \emph{Dual-Path Neck} with the pixel--volume 3DGS decoder.
Best results are highlighted in \textbf{bold}.}

\label{tab:abla}
\resizebox{1\linewidth}{!}{
\begin{tabular}{l|cccc}
\toprule
\textbf{Variant} & \textbf{PSNR} $\uparrow$ & \textbf{SSIM} $\uparrow$ & \textbf{LPIPS} $\downarrow$ & \textbf{PCC} $\uparrow$  \\
\midrule
SW1-None & 21.92 & 0.628 & 0.454 & 0.755 \\
SW2-Always & 24.01 & 0.738 & 0.252 & 0.814 \\
SW3-Half & 24.04 & 0.739 & 0.251 & 0.813 \\
\midrule
NH1-Geo-PV  & 20.47 & 0.557 & 0.527 & 0.619 \\
NH2-Geo-GS & 21.87 & 0.618 & 0.371 & 0.764 \\
NH3-Geo\&App-GS  & 23.83 & 0.730 & 0.271 & 0.681 \\
\midrule
\rowcolor{gray!10}\textbf{Ours: Full VGGD} & \textbf{24.12} & \textbf{0.741} & \textbf{0.247} & \textbf{0.814} \\
\bottomrule
\end{tabular}}
\end{table}